\documentclass[letterpaper, 10 pt, conference]{ieeeconf}  % Comment this line out if you need a4paper

\IEEEoverridecommandlockouts                              % This command is only needed if 
\usepackage{amsmath}
\usepackage{amsfonts}
\usepackage{dsfont}

\usepackage{booktabs} % tabels in the book booktabs style
\usepackage{multirow}
\usepackage{graphicx}

\usepackage{comment}
\usepackage{subcaption}
\usepackage{xcolor}

\usepackage{hyperref}
\usepackage{cleveref}

\usepackage{tikz}

\definecolor{bosch_red}{RGB}{226, 0, 20}
\definecolor{bosch_green}{RGB}{0, 98, 73}

\newcommand\copyrighttext{%
	\footnotesize \copyright\,2024 IEEE. Personal use of this material is permitted. Permission from IEEE must be obtained for all other uses, in any current or future media, including reprinting/republishing this material for advertising or promotional purposes, creating new collective works, for resale or redistribution to servers or lists, or reuse of any copyrighted component of this work in other works.}%

\newcommand\copyrightnotice{%
    \begin{tikzpicture}[remember picture,overlay]%
     \node[anchor=south, yshift=10pt] at (current page.south)%
     {\fbox{\parbox{\dimexpr\textwidth-\fboxsep-\fboxrule\relax}{\copyrighttext}}};%
     \end{tikzpicture}%
}

\begin{document}

\title{\LARGE \bf 
Revisiting Out-of-Distribution Detection in LiDAR-based 3D Object Detection
}

\hyphenation{Mi-cro-tech-no-lo-gy}
\author{Michael Kösel$^{1}$, Marcel Schreiber$^{2}$, Michael Ulrich$^{2}$, Claudius Gl\"aser$^{2}$ and Klaus Dietmayer$^{1}$
	\thanks{$^{1}$Institute of Measurement, Control, and Microtechnology, Ulm University, Germany, {\tt\small \{first.last\}@uni-ulm.de}}% <-this % stops a space
    \thanks{$^{2}$Robert Bosch GmbH, Corporate Research, 71272 Renningen, Germany, {\tt\small \{first.last\}@de.bosch.com} and {\tt\small michael.ulrich2@bosch.com}}%
}

\maketitle
\copyrightnotice
\thispagestyle{empty}
\pagestyle{empty}

\begin{abstract}

LiDAR-based 3D object detection has become an essential part of automated driving due to its ability to localize and classify objects precisely in 3D.
However, object detectors face a critical challenge when dealing with unknown foreground objects, particularly those that were not present in their original training data.
These out-of-distribution~(OOD) objects can lead to misclassifications, posing a significant risk to the safety and reliability of automated vehicles.
Currently, LiDAR-based OOD object detection has not been well studied.
We address this problem by generating synthetic training data for OOD objects by perturbing known object categories.
Our idea is that these synthetic OOD objects produce different responses in the feature map of an object detector compared to in-distribution~(ID) objects.
We then extract features using a pre-trained and fixed object detector and train a simple multilayer perceptron~(MLP) to classify each detection as either ID or OOD.
%In contrast to existing work, which resorts to unrealistic evaluation strategies, such as artificially inserting OOD objects into point clouds, our approach aims for a more realistic evaluation. 
In addition, we propose a new evaluation protocol that allows the use of existing datasets without modifying the point cloud, ensuring a more authentic evaluation of real-world scenarios. 
The effectiveness of our method is validated through experiments on the newly proposed nuScenes OOD benchmark. 
The source code is available at \url{https://github.com/uulm-mrm/mmood3d}.

\end{abstract}

\section{Introduction}

\begin{figure}[ht!]
    \centering
    \begin{subfigure}{0.48\textwidth}
        \centering
        \includegraphics[width=\textwidth]{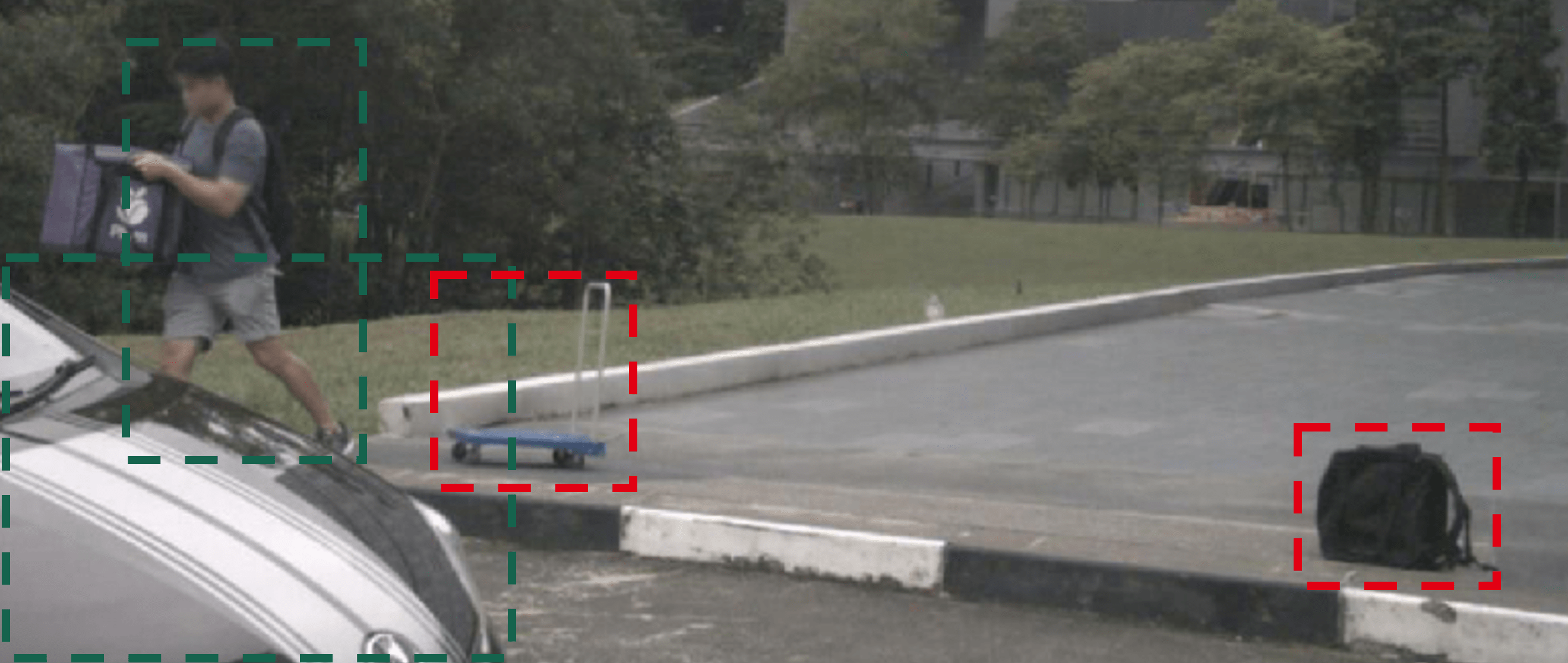}
    \end{subfigure}
    \begin{subfigure}{0.48\textwidth}
        \centering
        \includegraphics[width=\textwidth]{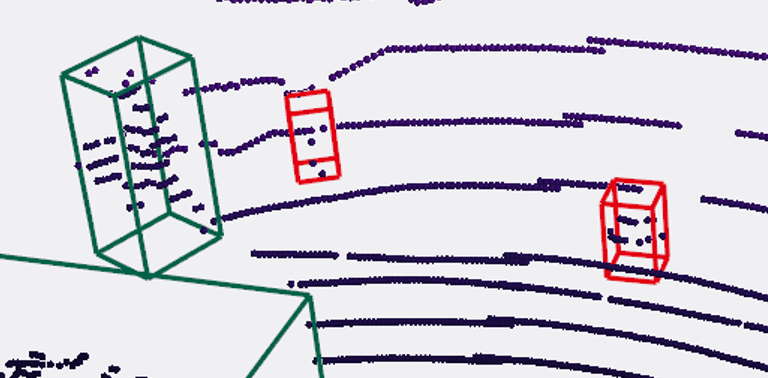}
    \end{subfigure}
    \caption{A scene containing a car, a pedestrian, and two unknown objects. ID objects are visualized in \textcolor{bosch_green}{green}, and OOD objects are visualized in \textcolor{bosch_red}{red}. Our proposed method is able to correctly classify the two unknown objects as OOD while still correctly classifying the known classes as ID. Camera image for visualization purposes only.}
    \label{fig:teaser}
\end{figure}

LiDAR-based 3D object detection has emerged as a fundamental technology in automated driving due to its ability to classify and localize objects in 3D~\cite{feng2020deep}.
Despite its advances, a critical challenge remains in the detection of out-of-distribution~(OOD) objects. 
These are object types that are not part of the training dataset of the object detectors.
Even though the detector is trained exclusively with in-distribution~(ID) data, it may still erroneously classify unknown objects as one of the ID classes with high confidence~\cite{nguyen2015}.
This challenge poses a significant threat to the safety and effectiveness of automated vehicles, as misclassifying unknown foreground objects can lead to severe misjudgments and safety risks.
An example of two OOD objects originally misclassified by the object detector but correctly classified by the OOD detector is shown in Fig.~\ref{fig:teaser}.
Recently, methods for OOD detection have evolved from the image classification task~\cite{MSP2016, odin2017, maxlogit2019, energy2020}, where the input images are classified as ID or OOD. 
The latest developments extend this to image-based object detection~\cite{du2022vos, Wilson_2023_ICCV, wu2023deep, kumar2023normalizing}, which aim to classify individual objects as either ID or OOD.
%Despite these advancements in the image domain, very little work has been done on OOD object detection in LiDAR point clouds.
Despite these advances in the image domain, very little work has been done on detecting OOD objects in LiDAR point clouds, where the challenges are increased by the complexity and sparsity of the data.
Huang~\textit{et al.}~\cite{huang2022out} introduced OOD detection to LiDAR-based 3D object detection.
Their method involves extracting features from an existing object detector and using them to train unsupervised learning methods, such as OC-SVM~\cite{scholkopf1999support} and normalizing flows~\cite{dinh2016density}.
Since there are currently no dedicated datasets for LiDAR-based OOD object detection in the context of autonomous driving, they synthetically create an evaluation dataset.
They insert objects, i.e., the corresponding LiDAR points from other real and simulated datasets, into the original point cloud, which they use for evaluation.
We argue that naively placing objects in another scene during evaluation does not necessarily guarantee a realistic evaluation of the OOD detection performance since this approach introduces a domain gap, such as differences in the intensities around the inserted objects or the lack of realistic occlusion patterns.

Our work addresses this gap and focuses on both improving OOD detection within LiDAR-based 3D object detection and introducing a realistic evaluation protocol for benchmarking the OOD methods in this field.
Inspired by~\cite{Wilson_2023_ICCV}, we propose a post-hoc OOD detection method, in which we extend a fixed 3D object detector with a multilayer perceptron~(MLP), which is trained to discriminate between features of ID and OOD objects.
This allows our method to be independent of the underlying object detector architecture.
We rely on the generation of synthetic data for OOD objects rather than requiring OOD objects from other datasets.
We use the existing annotated ID objects and modify them by randomly applying data augmentations to create synthetic OOD objects.
This allows us to train our auxiliary MLP using features extracted from both real ID and synthetic OOD objects.
Building on this, we propose a novel evaluation protocol specifically tailored for OOD object detection to address the limitation of not being able to accurately assess the performance in real-world scenarios.
This protocol is designed to better reflect the complexity of real-world applications, where scenes often contain both ID and OOD objects.
We apply this evaluation protocol to the large-scale nuScenes~\cite{caesar2020} dataset, allowing us to propose the nuScenes OOD benchmark.
Our experimental results, conducted on this benchmark, validate the effectiveness of our approach. 
We demonstrate superior performance in correctly classifying OOD objects compared to existing methods.
To summarize, the contributions of this paper are the following:
\begin{itemize}
    \item We propose to improve the OOD detection for LiDAR-based 3D object detection with a lightweight post-hoc approach that integrates information from both the feature space and the output space. 
    \item We present a novel evaluation protocol for OOD detection adapted to the object detection task, which better reflects the application in the real world.
    \item We evaluate our approach on our proposed nuScenes OOD benchmark and show significantly improved OOD detection performances compared to prior work.
\end{itemize}

\section{Related Work}
\subsection{LiDAR-based 3D Object Detection}

LiDAR-based 3D object detection aims to generate a set of 3D bounding boxes and their corresponding object category based on a LiDAR point cloud.
Two common approaches exist to extract features from point clouds: point-based and voxel-based approaches.
Point-based methods~\cite{qi2017pointnet, qi2017pointnet++} directly process the unordered point cloud without requiring a structured format.
Voxel-based methods such as VoxelNet~\cite{zhou2018voxelnet} first perform voxelization of the input point cloud and then extract features using 3D convolutions.
SECOND~\cite{yan2018second} improves on VoxelNet by using sparse convolutions, drastically reducing the runtime and the amount of required memory.
PointPillars~\cite{lang2019pointpillars} encodes the input point cloud into pillars, which are infinite height voxels, and uses 2D convolutions for feature extraction.
CenterPoint~\cite{yin2021center} predicts dense heatmaps of object centers.
Additional regression maps store the object attributes at the corresponding object centers.
This allows for the efficient extraction of objects without the need for anchor boxes.
We use the voxel-based CenterPoint as the base detector in our work due to its good object detection performance and simplicity.
%PV-RCNN~\cite{shi2020pv} and PV-RCNN++~\cite{shi2023pv} combine point-based and voxel-based feature extraction to achieve more accurate object detection.
%VoxelNeXt improves upon the idea of CenterPoint but directly predicts objects based on sparse voxel features, further reducing execution times.
Current 3D object detectors only focus on detecting ID objects.
In this work, we add an OOD detector to a pre-trained and fixed LiDAR object detector to additionally classify detections into ID and OOD.

\subsection{Out-of-Distribution Detection}

Out-of-distribution detection aims to detect samples that were not part of the training distribution.
Most OOD detection literature is concerned with deriving an OOD score from an already trained classifier.
Hendrycks and Gimpel~\cite{MSP2016} present a baseline for OOD detection, which is the maximum softmax probability~(MSP).
However, the MSP is known to be overconfident even for unseen samples~\cite{nguyen2015}.
ODIN~\cite{odin2017} improves the OOD detection by introducing temperature scaling into the softmax function and adding perturbations to the input.
MaxLogit~\cite{maxlogit2019} directly uses the raw output logits.
As a result, the value of the resulting OOD score is not limited to a certain range, allowing to find better separation thresholds.
Liu~\textit{et al.}~\cite{energy2020} introduce the energy score to differentiate between ID and OOD samples.
%Sun~\textit{et al.}~\cite{sun2021react} introduce ReAct based on the finding that batch normalization causes high activations even for OOD inputs.
%To mitigate this issue, they truncate activations above a certain threshold.
Another line of research uses OOD data to improve the detection of OOD samples.
Outlier exposure~\cite{hendrycks2018ood} tries to improve OOD detection using an auxiliary OOD dataset.
Hornauer and Belagiannis~\cite{hornauer2023heatmap} add a decoder to a pre-trained and fixed network. 
The decoder aims to produce heatmaps where ID images have zero response, while OOD images should produce a high response.

Du~\textit{et al.}~\cite{du2022vos} introduce OOD detection to image-based object detection.
Their method VOS generates outliers by treating the feature space as class-conditional normal distributions and sampling outlier features from low-likelihood regions.
This allows them to train an energy-based classifier that classifies objects into ID and OOD.
SAFE~\cite{Wilson_2023_ICCV} analyzes the effect of OOD objects on the different layers of the backbone of an object detector.
They find that backbone layers containing batch normalization and residual connections are extremely sensitive to OOD inputs.
By applying adversarial perturbations to the input image and extracting features from those sensitivity-aware feature~(SAFE) layers, they can train an MLP to differentiate between ID and OOD features.
Similar to SAFE, we also use an MLP to monitor the output of a pre-trained and frozen object detector.
Wu~\textit{et al.}~\cite{wu2023deep} use feature deblurring diffusion to generate features for OOD objects close to the classification boundary between ID and OOD objects.
Kumar~\textit{et al.}~\cite{kumar2023normalizing} improve upon the idea of VOS by learning a joint data distribution of all ID classes using invertible normalizing flow.
As a result, synthesized OOD objects have a lower likelihood, allowing for a better separation of ID and OOD objects.
The aforementioned image-based OOD object detection methods evaluate their method by considering all detections on the ID dataset as ID objects and all detections on the OOD dataset as OOD objects.
However, this evaluation strategy neglects the fact that in the real-world, both ID and OOD objects can coexist in the same scene.

While there is a lot of research in image-based OOD object detection, the LiDAR-based OOD object detection task is still barely addressed.
Huang~\textit{et al.}~\cite{huang2022out} were the first to formalize OOD detection in LiDAR-based 3D object detection.
They extract features from the PointPillars~\cite{lang2019pointpillars} object detector and compare different unsupervised methods, i.e., OC-SVM~\cite{scholkopf1999support}, Mahalanobis distance~\cite{lee2018simple}, and normalizing flows~\cite{dinh2016density}.
For the evaluation, simulated or real objects are pasted from different datasets into the point cloud.
However, placing objects from different datasets into the evaluation set poses a few issues, such as different intensities and missing occlusion behind the pasted object.
We propose a different evaluation strategy that eliminates the need to place objects in an existing dataset.
We consider rare classes as OOD, allowing us to directly use existing datasets.
Piroli~\textit{et al.}~\cite{piroli2023ls} transfer the idea of VOS to LiDAR-based object detection and generate virtual outliers using an auto-encoder.
They aim to accurately classify false object predictions as outliers, i.e., locations where no object should be predicted.
In contrast, our goal is to identify objects that should be predicted but are currently misclassified as one of the ID categories.

\section{Method}
\begin{figure*}[t]
\centering
\vspace{1pt}
  \includegraphics[width=1\textwidth]{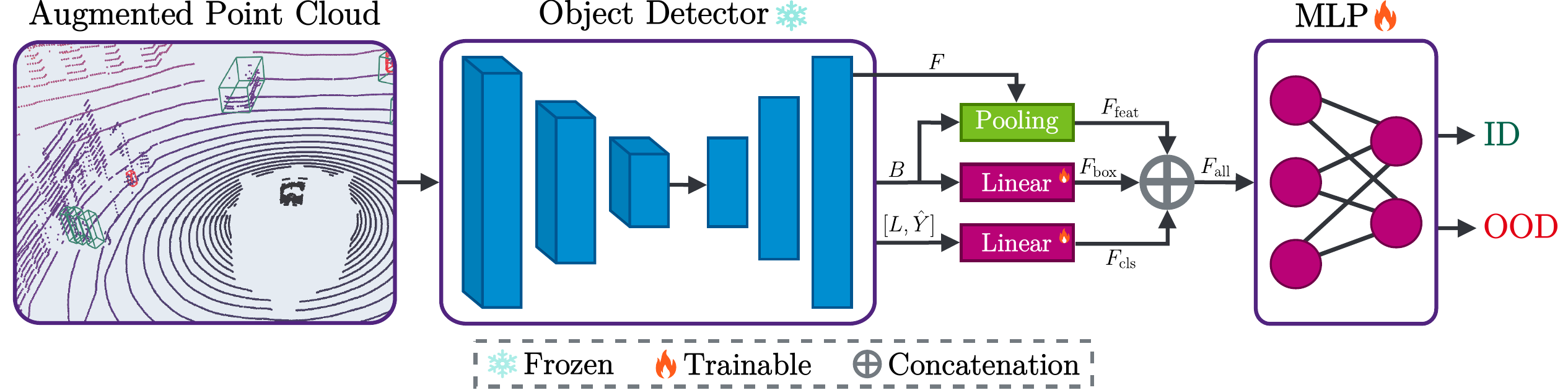}
  \caption{Overview of our proposed framework. We synthesize OOD objects by randomly selecting ID objects and scaling them by an unusual amount. This allows for the training of an MLP in a supervised manner. First, features $F_\text{feat}$ are extracted from the feature map $F$ of the object detector using the bounding boxes $B$. Then, the bounding boxes $B$ are encoded into $F_\text{box}$, the logits $L$ and the one-hot encoded classes $\hat{Y}$ are concatenated and encoded into $F_\text{cls}$. Finally, these features are concatenated with $F_\text{feat}$ to get $F_\text{all}$, which is used as input for the MLP. During training, the ground truth annotations are used instead of the predictions to obtain $B$, $L$, and $\hat{Y}$.}
  \label{fig:overview}
\end{figure*}
Our method involves implementing a post-hoc approach, i.e., using a pre-trained and fixed object detector, which removes the need for expensive retraining.
Synthetic OOD objects are generated by randomly augmenting annotated ground truth boxes of ID objects and the points within their bounding boxes.
A simple MLP is used for classification as a post-hoc method, which receives features from the detector's backbone and the detections themselves to classify into ID and OOD.
Note that the object detector is fixed, and only the MLP is optimized for the task of OOD detection.
An overview of our proposed method is shown in Fig.~\ref{fig:overview}.

\subsection{Problem Formulation}

The input to the 3D object detector is a point cloud, represented as $p \in \mathbb{R}^{N \times 4}$, where $N$ is the number of points in the cloud.
Typically, these $4$ features include the three spatial coordinates $(x, y, z)$ of each point in the point cloud, along with the intensity value of each point.
The 3D object detector $f(p)$ processes this point cloud to predict a set of bounding boxes $B = \{b_1, b_2, \dots, b_M\}$, where $M$ corresponds to the number of predicted boxes. 
Each bounding box $b_i \in \mathbb{R}^7$ is defined as $b_i = \{x_c, y_c, z_c, l, w, h, \theta\}$.
Here, $(x_c, y_c, z_c)$ denote the center coordinates of the box, $l$, $w$, and $h$ are the length, width, and height of the box, respectively, and $\theta$ is the rotation around the yaw axis.
Additionally, the object detector also outputs the raw logits $L = \{l_1, l_2, \dots, l_M\}$ with $l_i \in \mathbb{R}^K$ and the one-hot encoded predicted classes $\hat{Y} = \{\hat{y}_1, \hat{y}_2, \dots, \hat{y}_M\}$ with each predicted class $\hat{y}_i \in \mathbb{R}^K$.
For training, the detector uses a dataset containing only known classes, denoted $\mathcal{D}_{train}$. 
This dataset includes a finite set of classes $\mathcal{Y}_{train} = \{1, 2, \dots, K\}$, where $K$ represents the number of known classes. 
The detector is trained to localize and classify objects belonging to these ID classes.
During inference, the detector is confronted with known ID objects and never-before-seen OOD objects, and therefore OOD detection is required.
To evaluate the detector's performance on both ID and OOD objects, the detector is tested on a dataset $\mathcal{D}_{test}$, where each scene can contain both ID and OOD objects. 
The ID objects in $ \mathcal{D}_{test}$ belong to the classes in $\mathcal{Y}_{train}$, while all OOD objects are combined into a single class, denoted as $K+1$.
The goal of post-hoc OOD object detection is then to add the ability to detect OOD objects while maintaining the accuracy of classifying ID objects.

%The primary goal of OOD detection in LiDAR-based 3D object detection systems is to distinguish objects that do not belong to the known categories (ID) present in the training dataset. 

\subsection{Outlier Generation}
\label{sec:outlier_gen}

\begin{figure}[ht!]
    \centering
    \begin{subfigure}{0.236\textwidth}
        \centering
        \includegraphics[width=\textwidth]{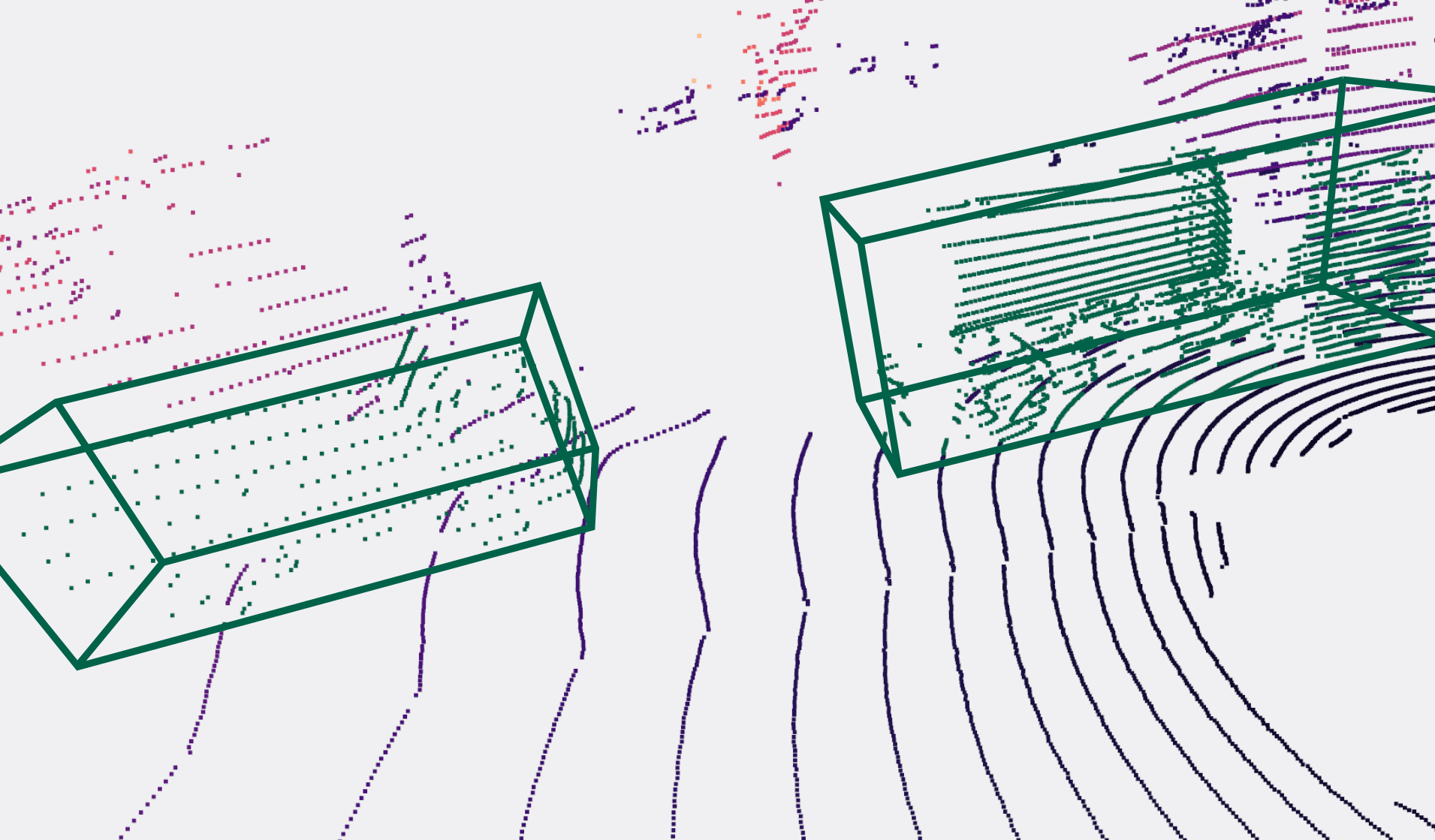}
        %\vspace{-0.45cm}
        \caption{Original}
        \label{fig:aug_orig}
    \end{subfigure}
%    \hspace{0.2em}
    \begin{subfigure}{0.236\textwidth}
        \centering
        \includegraphics[width=\textwidth]{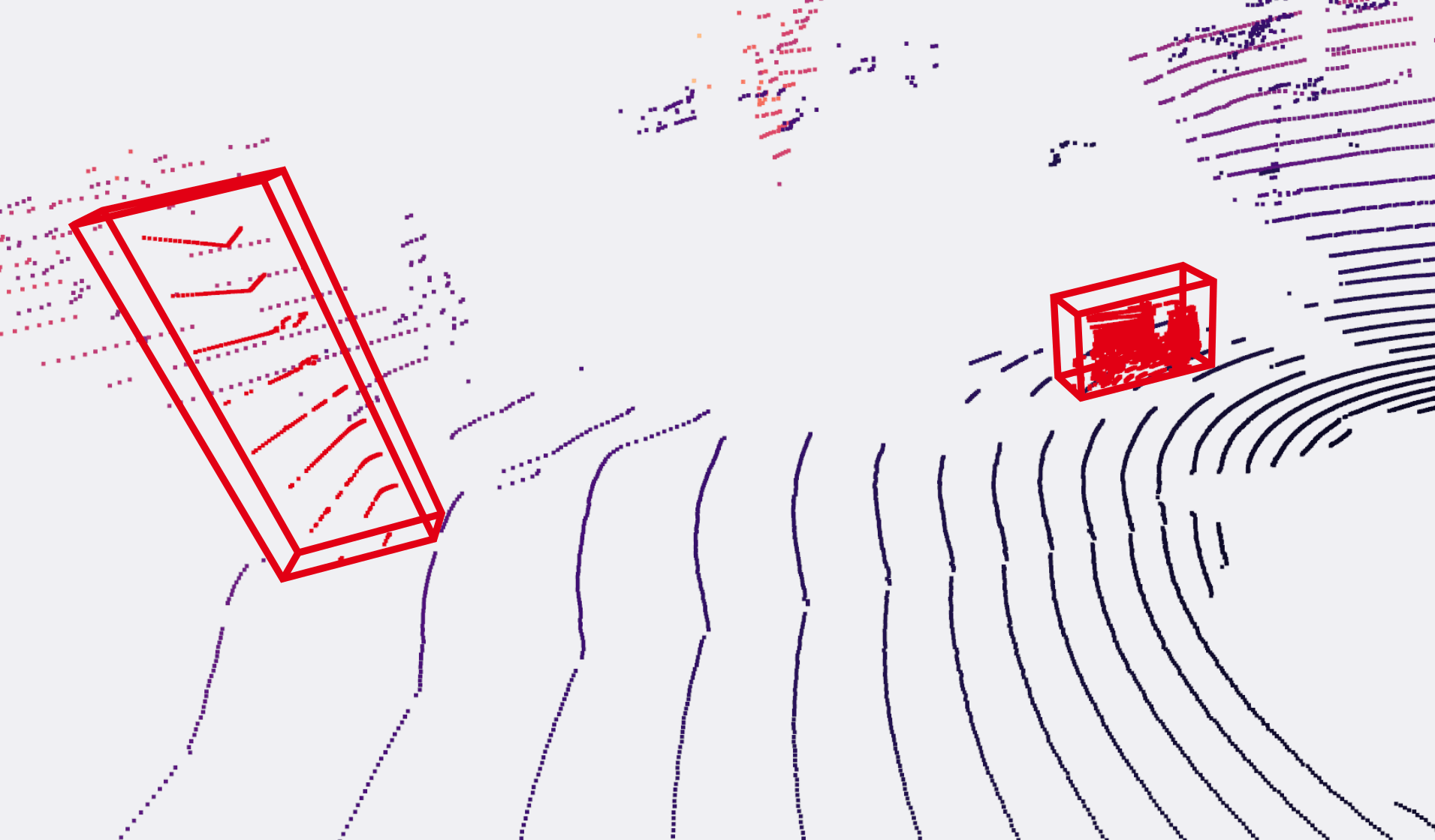}
        %\vspace{-0.45cm}
        \caption{Augmented}
        \label{fig:aug_shrinked}
    \end{subfigure}
    \caption{Example of the proposed OOD object synthesis. (a) two ID objects~(truck), (b) the resulting OOD objects}
    \label{fig:aug_vis}
\end{figure}

Acquiring real outliers is a time-consuming task, especially considering the vast number of different types of OOD objects.
To overcome this problem, we generate synthetic outliers based on the existing objects in the point cloud.
The synthetic outliers should share some geometric similarities with existing ID objects.
This ensures that the OOD objects do not share properties with static structures, such as buildings.
A simple but effective way to synthesize outliers is to randomly scale ID objects by an unusual amount, similar to \cite{cen2022openworld}.
The objects are scaled by factors that can either be very small or large, applied independently to each axis. 
This approach ensures that the resulting object does not simply resemble a slightly smaller or larger version of an ID object, which would complicate the classification process.
For each axis comprising width, height, and length, a random scaling factor is applied, varying from $0.1$ to $0.5$ for smaller adjustments and from $1.5$ to $3$ for larger adjustments.
For each axis, there is a $80\,\%$ probability of selecting a smaller scaling factor and a $20\,\%$ probability of selecting a larger one.
%There is an $80\,\%$ probability of selecting a smaller scaling factor and a $20\,\%$ probability for a larger one for each axis independently.
We empirically found that this method of independently scaling objects on different axes more frequently to smaller sizes yields better results than scaling them uniformly small or large. 
By using different random scaling factors for each axis, we create a more varied set of OOD objects.
The described scaling augmentation is applied to randomly selected ID objects in the scene.
First, we filter out objects containing less than $5$ points.
This ensures that the objects are representative enough to be used for learning OOD objects.
Then we randomly select $50\,\%$ of them, apply the scaling augmentation, and label them as OOD.
An example of our random scaling data augmentation is shown in Fig.~\ref{fig:aug_vis}.

\subsection{Feature Extraction}
\label{sec:feature_extraction}

Object detectors extract features from a point cloud to produce bounding boxes and class predictions for each object.
We aim to implement a feature extraction method independent of the underlying object detector architecture, allowing us to extract features corresponding to bounding boxes that can then be used for OOD detection.
Instead of expensive RoI-pooling~\cite{he2017mask} or RoI-align~\cite{girshick2014rich} operations, we simply extract features at a single location in the feature map.
Suppose we have a feature map $F \in \mathbb{R}^{H \times W \times C}$, where $H$ and $W$ are the height and width, respectively, and $C$ is the number of channels.
We extract features at the center position of each object along the channel dimension of $F$ using bilinear interpolation, resulting in $F_\text{feat} = \{F_\text{feat}^1, F_\text{feat}^2, \dots, F_\text{feat}^M\}$ where $F_\text{feat}^i \in \mathbb{R}^{C}$.
In addition to the feature maps, the pre-trained object detector provides other valuable information, such as the predicted boxes, categories, and classification logits.
Our strategy is to leverage this information to enhance OOD classification. 
Each bounding box $b_i$ is encoded using a linear layer:
\begin{equation}
\label{eq:box}
    F_\text{box}^i = \text{Linear}(b_i).
\end{equation}
Furthermore, we combine the raw output logits $l_i$ with the one-hot encoded class $\hat{y}$:
\begin{equation}
\label{eq:cls}
    F_\text{cls}^i = \text{Linear}([l_i, \hat{y}_i]),
\end{equation}
where $[\cdot]$ denotes a concatenation operation along the channel dimension.
For both $F_\text{box}$ and $F_\text{cls}$, the linear layer projects the features to a higher dimensional space, i.e., 64 dimensions.
This projection enriches the representational capacity of the features and allows the model to capture more complex patterns in the data.
Finally, the fused features $F_\text{all}$ are obtained by concatenating all features
\begin{equation}
\label{eq:fused}
    F_\text{all}^i = [F_\text{feat}^i, F_\text{box}^i, F_\text{cls}^i].
\end{equation}
The intuition behind this approach is to exploit discrepancies between predicted boxes, logits, and classifications to improve OOD detection. 
For instance, an inconsistency between the predicted class and the distribution of the output logits may indicate OOD samples. 
Similarly, anomalies in the predicted box dimensions, especially when they deviate significantly from the expected values for ID classes, e.g., an unusually large pedestrian, can signal OOD instances. 
By integrating this knowledge with the rich semantic information contained in the feature maps, we aim to improve the accuracy of the classifications.

\subsection{Feature Monitoring MLP}

The previous section explained how to extract features from a pre-trained object detector.
Our method is based on the assumption that OOD objects produce different features than known ID objects.
Using the features $F_\text{all}$, we want to train an MLP to discriminate between features of ID and OOD objects.
Similar to \cite{Wilson_2023_ICCV}, we use a simple 3-layer MLP.
The size of the input dimension is halved with each layer, and the output size is set to $1$. 
A dropout layer with a drop probability of $30\,\%$ is added before the final linear layer to prevent overfitting.
A Sigmoid activation is also applied to the output layer to produce a value between $[0, 1]$. 
A higher value indicates OOD objects, while a lower value indicates ID objects.

\subsection{Training and Inference}

To train the MLP, it is necessary to match the predictions of the detector to the ground truths.
However, the detector may be unable to localize synthetic OOD objects, particularly those scaled with a low or high factor, resulting in most of them not being detected.
Therefore, it is not possible to match the ground truth either because there is no corresponding prediction or because the predicted object's shape does not match the ground truth box well enough to be matched.
To address this issue, we extract the features $F_\text{feat}$ directly using the ground truth boxes during training instead of relying on the predicted boxes $B$.
Additionally, both $L$ and $\hat{Y}$ are obtained using the ground truths during training instead of the predicted outputs.
We optimize using the binary cross-entropy~(BCE) loss, where OOD objects are labeled $1$, and ID objects are labeled $0$.
Therefore, the MLP is trained to output higher values for OOD objects and lower values for ID objects.

During inference, the predictions of the base detector are used for feature extraction.
Given the predicted score $g(x)$ of the MLP, the binary classification rule can then be formulated as follows
\begin{equation}
\label{eq:decision_rule}
%\hat{y}_o= \begin{cases} \text{ID} & \text { if } g(x) \leq \delta \\ \text{OOD} & \text { otherwise }
\hat{y}_o= \begin{cases} 0 & \text { if } g(x) \leq \delta \\ 1 & \text { otherwise }
\end{cases},
\end{equation}
where $\delta$ is a threshold that determines what is classified as ID or OOD.
In practice, the threshold $\delta$ is chosen such that a high proportion of the ID objects are correctly classified~(e.g., $95\,\%$).

\subsection{Evaluation Protocol}
\label{sec:eval_protocol}

Current image-based OOD object detection methods~\cite{du2022vos, Wilson_2023_ICCV, wu2023deep, kumar2023normalizing}, employ an evaluation protocol that is derived from the approach used in OOD classification literature~\cite{MSP2016, odin2017, maxlogit2019, energy2020}.
This protocol involves using separate datasets for ID and OOD samples, with no overlap between the classes of both datasets.
Each detection in the ID dataset is then treated as an ID sample, while each object detection in the OOD dataset is considered OOD.
However, applying this approach to object detection neglects the fact that scenes can simultaneously contain both ID and OOD objects.

Current literature~\cite{huang2022out} in LiDAR-based OOD object detection utilizes a more realistic strategy of considering scenes that contain both ID and OOD objects.
To construct such scenes, objects from simulators such as CARLA~\cite{dosovitskiy2017carla} or from different datasets are placed into the evaluation scenes.
While this approach allows for a more realistic evaluation by including both ID and OOD objects in the same scenes, it introduces additional concerns.
Firstly, simply placing objects in the point cloud disregards the LiDAR's scanning principle, i.e., no occlusion behind the inserted objects.
Secondly, the intensity values of the inserted objects differ from those of other points in the scene.
The evaluated object detectors could use these clues to aid in classification, meaning their performance in real-world scenarios cannot be reliably assessed.
Note that the outlier synthesis described in Sec.~\ref{sec:outlier_gen} produces objects unlike real LiDAR data; however, this only impacts training and not evaluation scenes.

Current automotive LiDAR datasets contain annotations for common objects, such as cars, pedestrians, and bicycles.
However, they also often include other rare object categories where fewer annotations exist.
Due to the absence of annotations, these classes are frequently disregarded in evaluation benchmarks because of insufficient training data.
We propose to consider these rare object categories as OOD classes.
As rare objects in the dataset, they fulfill the property of outliers in real scenes.
To ensure that these objects are considered OOD, we remove any scenes containing at least one instance of an OOD object, as training data may contain scenes with both ID and OOD objects.
Furthermore, this ensures that the detector is not specifically trained to classify these objects as background, which could result in missed detections during evaluation.
This straightforward approach enables a realistic and straightforward evaluation of OOD object detectors, allowing us to use the provided test set directly by treating the rare classes as OOD objects.

\section{Experiments}

In this section, we present a quantitative evaluation of our approach in comparison to existing work on OOD detection. 
Additionally, we provide qualitative results of our method.
We use the well-established CenterPoint~\cite{yin2021center} object detector for all our experiments.

\subsection{Dataset}

We use the large-scale automated driving dataset nuScenes~\cite{caesar2020}.
nuScenes was recorded in Boston and Singapore.
The dataset consists of 1000 driving scenes, each approximately $20$\,s in duration, captured using a Velodyne HDL-32E 32-beam LiDAR sensor operating at a frequency of $20$\,Hz.
The driving sequences are split into $700$ training, $150$ validation, and $150$ testing scenes.
nuScenes uses the mean average precision~(mAP) and the nuScenes detection score~(NDS) as the primary main 3D object detection metrics.
Our evaluation does not utilize the official testing set as it lacks the ground truth data required for our task. 

Based on the new evaluation protocol of Section~\ref{sec:eval_protocol}, we propose the nuScenes OOD benchmark.
For our benchmark, we consider the \textit{void}-classes in the official detection benchmark as OOD.
The classes we use for our nuScenes OOD benchmark and their class distribution in the validation set are given in Table~\ref{tab:class_dist}.
\begin{table}[t!]
    \centering
    \caption{Distribution of ID and OOD objects in the nuScenes validation set.}
    \label{tab:class_dist}
    \setlength{\tabcolsep}{12pt} % default value is 6pt
    %\resizebox{0.98\columnwidth}{!}{%
    \begin{tabular}{@{}lc|lc@{}}
%    \vspace{5pt}
        \toprule
        \multicolumn{2}{c|}{\textbf{ID Objects}} & \multicolumn{2}{c}{\textbf{OOD Objects}} \\ \midrule
        \textbf{Category} & \textbf{\#} & \textbf{Category} & \textbf{\#} \\ \midrule
        car & 80004 & animal & 37 \\
        truck & 15704 & debris & 620 \\
        construction vehicle & 2678 & pushable-pullable & 3332 \\
        bus & 3158 & personal mobility & 16 \\
        trailer & 4159 & stroller & 131 \\
        barrier & 26992 & wheelchair & - \\
        motorcycle & 2508 & bicycle rack & 274 \\
        bicycle & 2381 & ambulance vehicle & 30 \\
        pedestrian & 34347 & police vehicle & 73 \\
        traffic cone & 15597 & & \\ 
        \midrule
        \textbf{Total} & 204528 & \textbf{Total} & 4513 \\ 
        \bottomrule
    \end{tabular}
    %}
\end{table}

\subsection{Evaluation Metrics}

We evaluate our method using standard OOD metrics~\cite{MSP2016, odin2017, maxlogit2019, energy2020, huang2022out}, specifically FPR at 95\% TPR~(FPR-95), AUROC, AUPR-Success~(AUPR-S), and AUPR-Error~(AUPR-E).
These metrics are computed independently of the OOD detection threshold $\delta$, enabling their calculation without pre-setting an appropriate threshold.
The FPR-95 metric measures the false positive rate~(FPR) when the true positive rate~(TPR) is at $95\%$, signifying the method's accuracy in correctly classifying an ID object as such.
Moreover, the area under the receiver operating characteristic~(AUROC) is crucial for evaluating the discriminative performance, with a value of $50\,\%$ indicating performance equivalent to random guessing.
Furthermore, we employ both variants of the area under the precision-recall curve~(AUPR): AUPR-Success and AUPR-Error. 
AUPR is particularly useful in scenarios with class imbalances~(see Table~\ref{tab:class_dist}).
AUPR-Success is calculated when ID is treated as the positive class, while AUPR-Error is computed when OOD is considered the positive class.
This differentiation helps to understand the model's ability to distinguish between ID and OOD objects.

In order to calculate the metric, it is necessary to separate the data since each scene can contain both ID and OOD objects. 
To this end, predictions are assigned to ground truths if their bird's-eye view~(BEV) center distance is $< 0.5$\,m, similar to the calculation of the mAP metric in nuScenes.
Then, each detection can be split into ID and OOD based on the assigned ground truth label, without requiring the use of the classification rule of Eq.~\ref{eq:decision_rule}, which requires an appropriate threshold.
Note that the OOD metrics are calculated for all predictions, which can be matched to ground truths. 
Since all investigated methods are post-hoc, each is evaluated with the same set of predictions.

\subsection{Baseline Methods}
We compare our proposed method with existing post-hoc methods in OOD classification and LiDAR-based OOD object detection.
For the baseline classification methods, we use MSP~\cite{MSP2016}, ODIN~\cite{odin2017}, MaxLogit~\cite{maxlogit2019}, and Energy~\cite{energy2020}.
Specifically for ODIN, we set the temperature parameter to $1000$ and do not use any input perturbations. 
The energy score is calculated using the default temperature setting of $1$.
All of the classification methods above require the raw output logits.
Since CenterPoint~\cite{yin2021center} outputs multiple heatmap heads, where each head groups similar classes, we concatenate the group-specific heatmaps into a single heatmap. 
Note that we use the raw heatmaps before applying the Sigmoid function.
This allows for the extraction of logits at each predicted location in a consistent manner for the application of the OOD methods.
Additionally, we include a comparison to the normalizing flow method used in \cite{huang2022out}.
We replicate this normalizing flow-based approach with RealNVP~\cite{dinh2016density}, using $32$ normalizing flow layers and a hidden dimension of $1024$.
In line with \cite{huang2022out}, features are extracted from the neck feature map.
Finally, we also compare our method to the predicted confidence score of the CenterPoint detector, which will be referred to as the default score in the following.
Given the random nature of learning-based methods, we report the mean and standard deviation over five runs with different seeds for these methods.

\subsection{Implementation Details}

We implement our method using PyTorch~\cite{paszke2019pytorch} in the MMDetection3D~\cite{mmdet3d2020} framework.
We choose the one-stage CenterPoint-Voxel object detector as the base detector for all experiments.
The CenterPoint detector is trained with the default settings of MMDetection3D on $8$ RTX 2080Ti GPUs with an effective batch size of $32$.
%We choose the voxel-based CenterPoint with a voxel size of $(0.1, 0.1, 0.2)$.
Following the evaluation protocol of Section~
\ref{sec:eval_protocol}, scenes from the nuScenes dataset that contain OOD objects are excluded when training the base detector.
The official CenterPoint detector achieves a mAP of $55.61$ and an NDS of $64.13$.
Our version, trained without frames containing OOD objects, achieves an mAP of $54.06$ and an NDS of $62.90$, which is reasonable given that approximately $35\,\%$ of the training data was removed.

To train our proposed method, we extract features from the neck feature map, which contains $512$ channels.
We train the MLP for $5$ epochs using the stochastic gradient descent~(SGD)~\cite{sutskever2013importance} optimizer.
The initial learning rate is set to $10^{-3}$, with a momentum of $0.9$, a weight decay of $10^{-4}$, and an effective batch size of $16$. 
We use a poly learning rate scheduler with a power setting of $3$ and a minimal learning rate of $10^{-5}$. 
This aggressive learning rate strategy minimizes the risk of overfitting, especially on the synthetic OOD objects introduced during training.
We use the same data augmentations as in the training of the base detector but remove the GT-sampling~\cite{yan2018second} augmentation since we have observed degraded performance when using it.
Our method adds only around $2$\,ms of overhead to the base detector.

\subsection{Quantitative Results}

We compare our methods against existing methods in OOD classification and detection.
The results on the nuScenes validation set are shown in Table~\ref{tab:results_val}.
\begin{table}[tb!]
    \centering
    \caption{OOD detection results on our proposed nuScenes OOD benchmark.}
    \label{tab:results_val}
    \setlength{\tabcolsep}{5pt} % default value is 6pt
    %\resizebox{0.98\columnwidth}{!}{%
    \begin{tabular}{@{}l|ccccc}%@{}}
        \toprule
        \textbf{Method}   & \textbf{FPR-95~$\downarrow$}   & \textbf{AUROC~$\uparrow$} & \textbf{AUPR-S~$\uparrow$} & \textbf{AUPR-E~$\uparrow$} \\ \midrule
        Default Score      & 63.00 & 85.10 & 99.70 & 7.91 \\
        MSP~\cite{MSP2016}      & 44.60 & 88.87 & \textbf{99.79} & 13.74 \\
        ODIN~\cite{odin2017}     & 48.63 & 87.82 & 99.77 & 11.05 \\
        MaxLogit~\cite{maxlogit2019} & 55.85 & 85.89 & 99.71 & 10.02 \\
        Energy~\cite{energy2020}   & 73.80 & 82.15 & 99.64 & 5.27 \\
        \midrule
        NormFlows~\cite{huang2022out} & 85.88$\pm$8.0 & 66.98$\pm$13.4 & 99.25$\pm$0.4 & 2.98$\pm$1.1 \\
        \midrule
        Ours & \textbf{36.96}$\pm$0.2 & \textbf{88.96}$\pm$0.1 & 99.73$\pm$0.0 & \textbf{24.68}$\pm$0.7 \\ \bottomrule
    \end{tabular}
   % }
\end{table}
Our method improves upon the other best-performing method in our evaluation, MSP, by $7.64\,\%$ in the FPR-95 metric.
The significant improvement of $10.94\,\%$ in the \mbox{AUPR-E} metric suggests that our method is more effective in accurately identifying OOD objects than existing methods.
The slightly lower results in the \mbox{AUPR-S} metric can be attributed to our synthetic OOD object generation method, which utilizes ID objects.
We randomly scale ID objects to generate ground truth data for OOD objects.
In some cases, a vehicle may be scaled to the point where it resembles a different type of vehicle, such as a car appearing to be a truck or vice versa.
This ambiguity can cause the classifier to incorrectly classify ID objects as OOD.
In our setup, the normalizing flows~\cite{huang2022out} method achieves the worst results.

\subsection{Qualitative Results}
\begin{figure*}[ht!]
    \centering
    \vspace{2pt}
    \begin{subfigure}{0.326\textwidth}
        \centering
        \includegraphics[width=\textwidth]{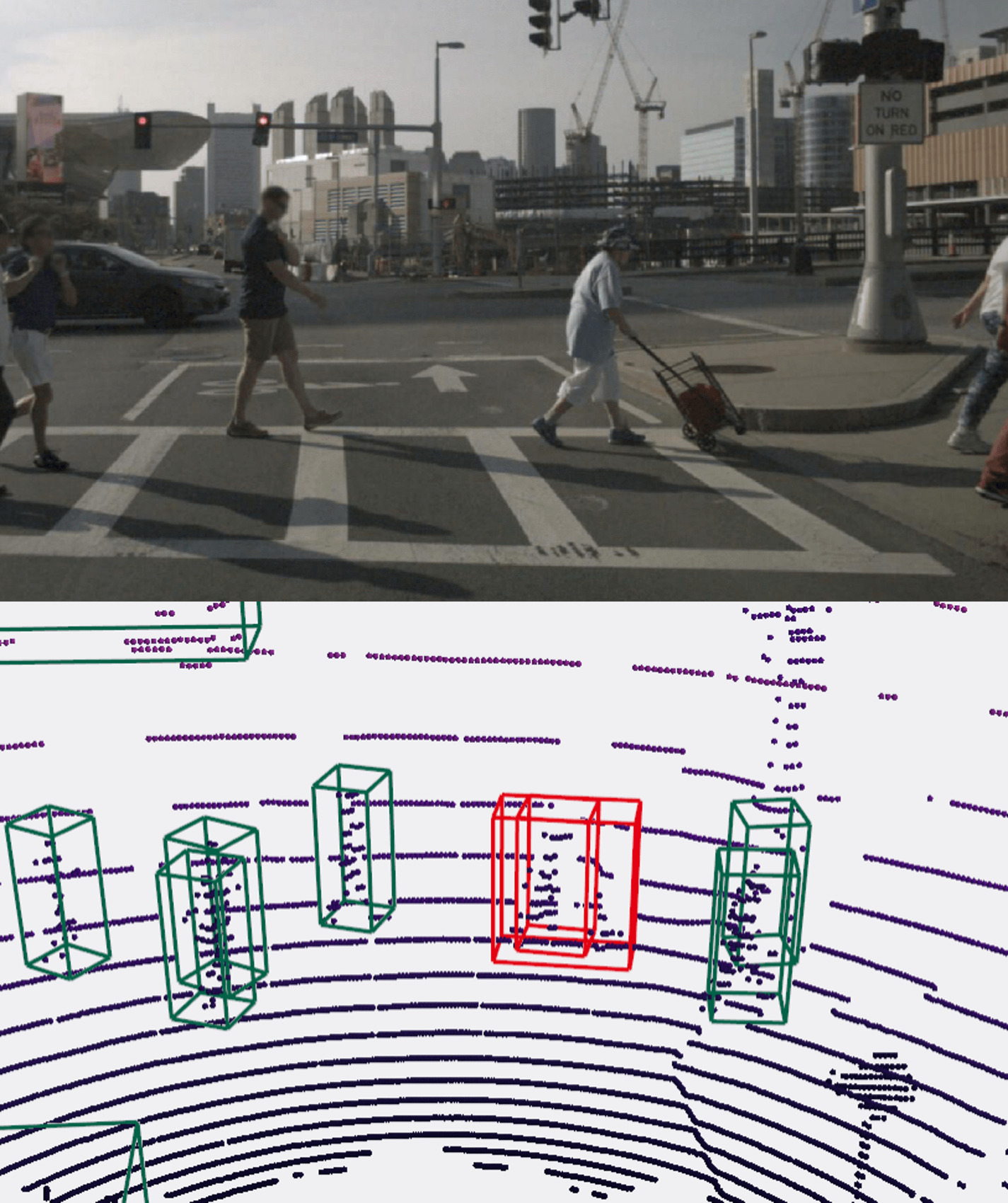}
        \caption{pushable-pullable}
        \label{fig:qual_pushable_pullable}
    \end{subfigure}
%    \hspace{0.2em}
    \begin{subfigure}{0.326\textwidth}
        \centering
        \includegraphics[width=\textwidth]{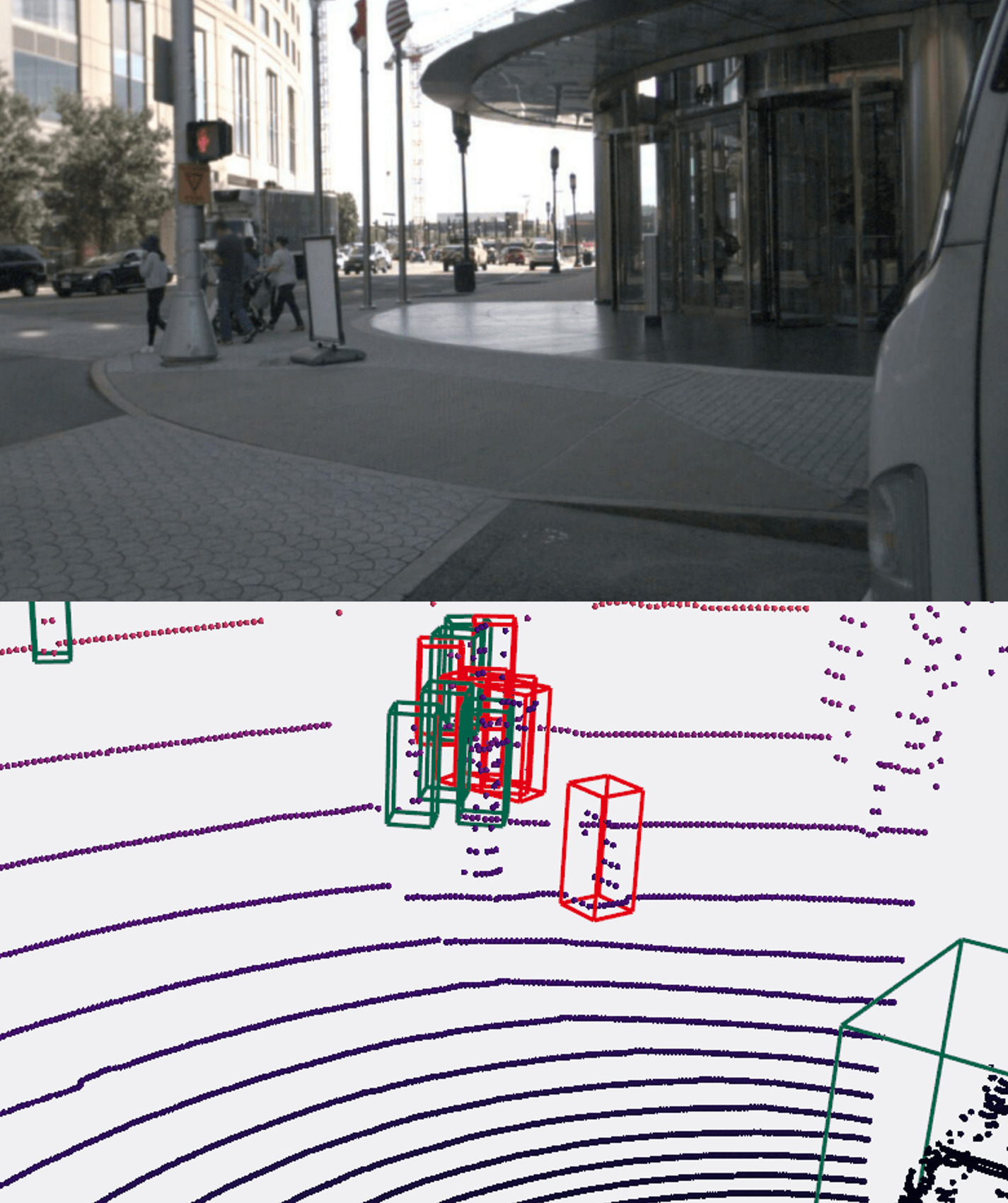}
        \caption{stroller}
        \label{fig:qual_stroller}
    \end{subfigure}
    \begin{subfigure}{0.326\textwidth}
        \centering
        \includegraphics[width=\textwidth]{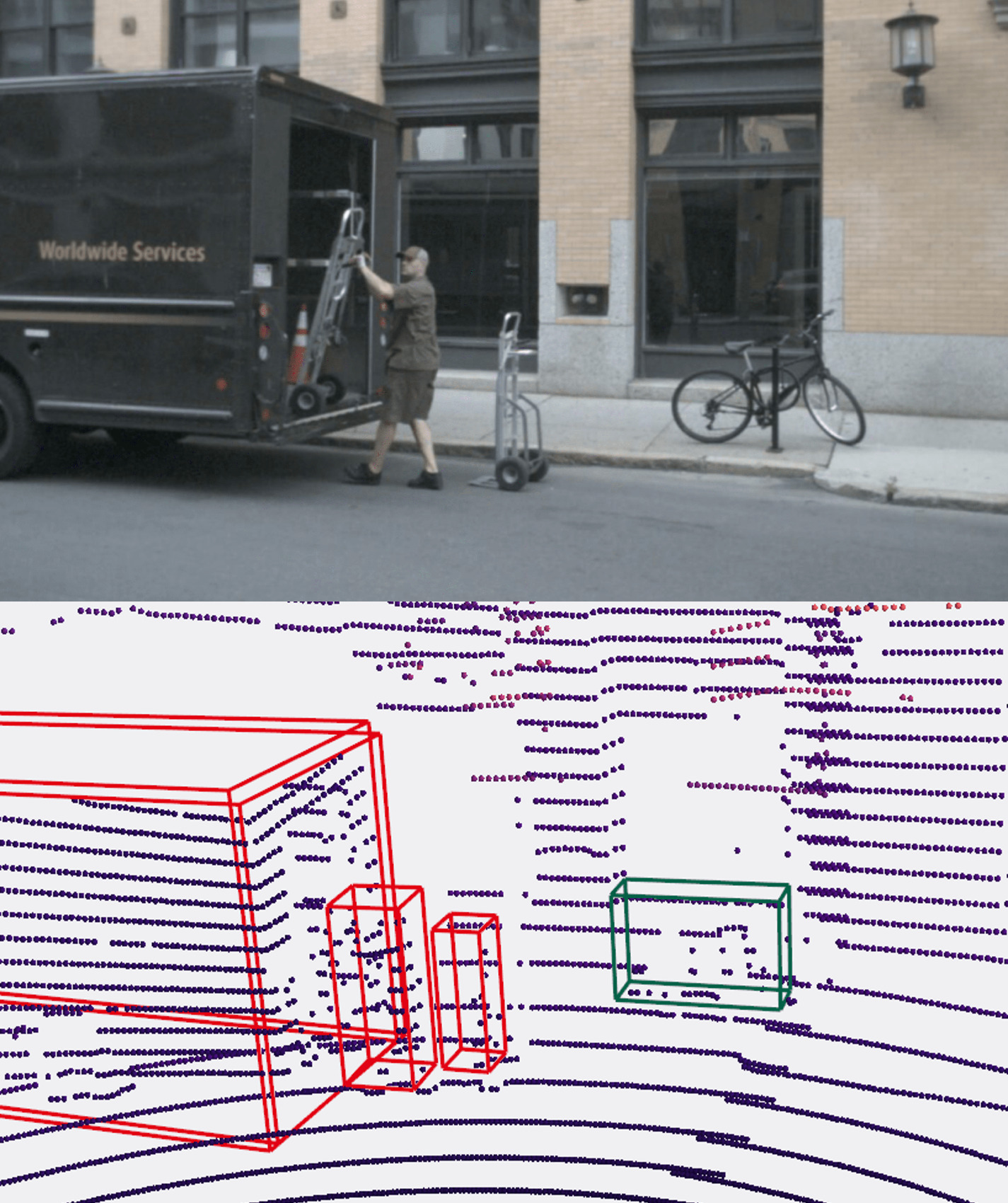}
%        \includegraphics[width=\textwidth]{imgs/qual_debris-min.png}
%        \caption{debris}
        \caption{pushable-pullable}
        \label{fig:qual_debris}
    \end{subfigure}
    \caption{Qualitative results of the OOD object detection on the nuScenes validation set. ID objects are visualized in \textcolor{bosch_green}{green} and OOD objects in \textcolor{bosch_red}{red}. The OOD classification threshold is chosen such that the true positive rate is $95\,\%$. Camera image for visualization purposes only.}
    \label{fig:qual_vis}
\end{figure*}

In Fig.~\ref{fig:qual_vis}, we show qualitative results on the nuScenes validation set.
Our method is able to correctly classify the corresponding OOD objects.
Fig~\ref{fig:qual_pushable_pullable} shows a pedestrian holding a pushable-pullable object. The CenterPoint detector initially detected a bicycle and a pedestrian.
Our OOD detector correctly classifies this as OOD.
In Fig~\ref{fig:qual_stroller}, there is a group of pedestrians, with the stroller being the OOD object of interest.
The stroller is detected by the original detector as a bicycle and several pedestrians.
The OOD detector correctly classifies the stroller as OOD and additionally classifies the redundant boxes as OOD.
In addition, the scene contains a signboard that is not part of the ID classes and not part of the evaluated OOD classes, hence it is considered a false positive object prediction. 
However, our method correctly identifies it as an OOD object instead of a pedestrian.
Finally, Fig~\ref{fig:qual_debris} depicts two pushable-pullable OOD objects that are correctly classified as OOD.
However, the OOD detector incorrectly classifies the mail delivery truck on the left as OOD.
We argue that the OOD classification can still be considered sound, given the difference in appearance between this type of vehicle and other trucks.

\subsection{Ablation Studies}

In the following, we analyze the influence of our design choices on the final performance.

\subsubsection{Choice of feature maps}

The following is an analysis of which feature maps are most suitable for feature extraction.
In particular, the backbone of the voxel-based CenterPoint detector consists of: (1) four sparse 3D encoder layers, (2) one 2D spatial layer, (3) two 2D backbone layers, and (4) one neck layer.
Since the layers of (1) are 3D, we project them to the BEV by taking the sum along the depth dimension.
Additionally, we want to analyze whether or not SAFE layers~\cite{Wilson_2023_ICCV} provide improved discrimination capabilities for our method.
The layers referred to as SAFE layers contain both a batch normalization and a residual connection.
In our case, only the layers of (1) fulfill the property of a SAFE layer.
The performance of the different feature maps is given in Table~\ref{tab:ablation_feat_map}.
\begin{table}[tb!]
\centering
\caption{Effect of the choice of feature map(s) on the final results.}
\label{tab:ablation_feat_map}
\setlength{\tabcolsep}{5.5pt} % default value is 6pt
%\resizebox{1\columnwidth}{!}{%
\begin{tabular}{@{}c|ccccc@{}}
\toprule
\textbf{Feature Maps} & \textbf{FPR-95~$\downarrow$}   & \textbf{AUROC~$\uparrow$} & \textbf{AUPR-S~$\uparrow$} & \textbf{AUPR-E~$\uparrow$} \\ \midrule
(1) & 56.37$\pm$1.0 & 84.85$\pm$0.4 & 99.68$\pm$0.0 & 9.67$\pm$0.2 \\
(2) & 46.40$\pm$0.8 & 87.41$\pm$0.1 & 99.72$\pm$0.0 & 12.29$\pm$0.2 \\
(3) & 39.09$\pm$0.3 & 88.94$\pm$0.1 & 99.72$\pm$0.0 & 19.71$\pm$0.5 \\
%\midrule
(4) & \textbf{36.96}$\pm$0.2 & \textbf{88.96}$\pm$0.1 & \textbf{99.73}$\pm$0.0 & \textbf{24.68}$\pm$0.7 \\ \bottomrule
\end{tabular}
%}
\end{table}
We can see that, in our case, the performance of the SAFE layers is significantly worse than that of the other later layers.
We attribute this to the fact that the later layers contain more aggregated semantic information, which allows for better performance in distinguishing between features of ID objects and features of OOD objects.
This is also consistent with the results of Huang~\textit{et al.}~\cite{huang2022out}, where later features proved to be more performant.

\subsubsection{Feature fusion}
We also compare the effect of fusing the object detector predictions with features extracted from the feature map.
The findings of this study are presented in Table~\ref{tab:fusion_ablation}.
\begin{table}[tb!]
\centering
\caption{Impact of feature fusion on the final OOD detection results.}
\label{tab:fusion_ablation}
\setlength{\tabcolsep}{4.6pt} % default value is 6pt
%\small
%\resizebox{0.98\columnwidth}{!}{%
\begin{tabular}{@{}ccc|cccc}%@{}}
\toprule
$F_\text{feat}$ & $F_\text{box}$ & $F_\text{cls}$ & \textbf{FPR-95~$\downarrow$}   & \textbf{AUROC~$\uparrow$} & \textbf{AUPR-S~$\uparrow$} & \textbf{AUPR-E~$\uparrow$} \\
\midrule
% \hline
$\checkmark$ &  &  & 39.51$\pm$0.5 & 88.58$\pm$0.0 & 99.72$\pm$0.0 & 20.82$\pm$0.3 \\
$\checkmark$ & $\checkmark$ & & 38.97$\pm$0.4 & 88.56$\pm$0.1 & \textbf{99.73}$\pm$0.0 & 20.54$\pm$0.4 \\
$\checkmark$ & & $\checkmark$ & 37.97$\pm$0.6 & 88.84$\pm$0.1 & \textbf{99.73}$\pm$0.0 & 23.88$\pm$0.7 \\
$\checkmark$ & $\checkmark$ & $\checkmark$ & \textbf{36.96}$\pm$0.2 & \textbf{88.96}$\pm$0.1 & \textbf{99.73}$\pm$0.0 & \textbf{24.68}$\pm$0.7 \\
\bottomrule
\end{tabular}%
%}
\end{table}
The use of only $F_\text{feat}$ results in less effective results, mainly due to the lack of additional information about the predictions. 
However, when $F_\text{cls}$ is incorporated, there is a notable improvement in results, surpassing those achieved with $F_\text{box}$. 
This is likely due to the valuable insights from the discrepancies between predicted logits and the predicted classes. 
Integrating all three features, i.e., $F_\text{feat}$, $F_\text{cls}$, and $F_\text{box}$ further improves the overall performance.

\subsubsection{OOD Synthesis}

In Table~\ref{tab:augmentation_ablation}, we show the difference between having an independent scaling factor for each axis and having the same scaling factor for each axis.
\begin{table}[tb!]
    \centering
    \caption{Analysis of the random scaling augmentations used for OOD object synthesis.}
    \label{tab:augmentation_ablation}
    \setlength{\tabcolsep}{5pt} % default value is 6pt
    %\resizebox{0.98\columnwidth}{!}{%
    \begin{tabular}{@{}l|ccccc@{}}
        \toprule
        \textbf{Random Scaling}   & \textbf{FPR-95~$\downarrow$}   & \textbf{AUROC~$\uparrow$} & \textbf{AUPR-S~$\uparrow$} & \textbf{AUPR-E~$\uparrow$} \\ \midrule
        Equal            & 38.99$\pm$0.2 & 88.92$\pm$0.1 & \textbf{99.75}$\pm$0.0 & 20.73$\pm$0.8 \\
        Independent      & \textbf{36.96}$\pm$0.2 & \textbf{88.96}$\pm$0.1 & 99.73$\pm$0.0 & \textbf{24.68}$\pm$0.7 \\
        \bottomrule
    \end{tabular}
    %}
\end{table}
Applying random scaling independently to each axis produces better results than equal scaling, as it allows for more variation in the synthesized objects, making them more distinguishable from ID objects.

\section{Conclusion}
This paper presents a method for accurately identifying OOD objects in real-world scenarios.
In order to generate training data for OOD objects, we apply random scaling to randomly selected ID objects in the input point cloud and use a simple MLP for the OOD classification.
Our method has the advantage of not requiring expensive base detector retraining but only training a simple MLP.
Consequently, our method makes no assumptions about the underlying object detector architecture.
For the evaluation of our and other methods, we propose a novel evaluation protocol where we consider rare classes as OOD objects.
We applied this evaluation protocol on the large-scale autonomous driving dataset nuScenes and thereby introduce the nuScenes OOD benchmark.
This approach is more realistic for assessing performance in the real world than existing methods that synthetically place OOD objects in point clouds during evaluation.
We evaluate our method against existing methods on the proposed nuScenes OOD benchmark and achieve superior results. 


%%%%%%%%%%%%%%%%%%%%%%%%%%%%%%%%%%%%%%%%%%%%%%%%%%%%%%%%%%%%%%%%%%%%%%%%%%%%%%%%

%%%%%%%%%%%%%%%%%%%%%%%%%%%%%%%%%%%%%%%%%%%%%%%%%%%%%%%%%%%%%%%%%%%%%%%%%%%%%%%%

%%%%%%%%%%%%%%%%%%%%%%%%%%%%%%%%%%%%%%%%%%%%%%%%%%%%%%%%%%%%%%%%%%%%%%%%%%%%%%%%

\bibliographystyle{IEEEtran}
\bibliography{library}

\begin{thebibliography}{10}
\providecommand{\url}[1]{#1}
\csname url@samestyle\endcsname
\providecommand{\newblock}{\relax}
\providecommand{\bibinfo}[2]{#2}
\providecommand{\BIBentrySTDinterwordspacing}{\spaceskip=0pt\relax}
\providecommand{\BIBentryALTinterwordstretchfactor}{4}
\providecommand{\BIBentryALTinterwordspacing}{\spaceskip=\fontdimen2\font plus
\BIBentryALTinterwordstretchfactor\fontdimen3\font minus
  \fontdimen4\font\relax}
\providecommand{\BIBforeignlanguage}[2]{{%
\expandafter\ifx\csname l@#1\endcsname\relax
\typeout{** WARNING: IEEEtran.bst: No hyphenation pattern has been}%
\typeout{** loaded for the language `#1'. Using the pattern for}%
\typeout{** the default language instead.}%
\else
\language=\csname l@#1\endcsname
\fi
#2}}
\providecommand{\BIBdecl}{\relax}
\BIBdecl

\bibitem{feng2020deep}
D.~Feng, C.~Haase-Schütz, L.~Rosenbaum, H.~Hertlein, C.~Gläser, F.~Timm,
  W.~Wiesbeck, and K.~Dietmayer, ``{Deep Multi-Modal Object Detection and
  Semantic Segmentation for Autonomous Driving: Datasets, Methods, and
  Challenges},'' \emph{IEEE Trans. Intell. Transp. Syst.}, vol.~22, no.~3, pp.
  1341--1360, 2021.

\bibitem{nguyen2015}
A.~Nguyen, J.~Yosinski, and J.~Clune, ``{Deep Neural Networks are Easily
  Fooled: High Confidence Predictions for Unrecognizable Images},'' in
  \emph{Proc. IEEE Conf. Comput. Vis. Pattern Recog.}, 2015, pp. 427--436.

\bibitem{MSP2016}
D.~Hendrycks and K.~Gimpel, ``{A Baseline for Detecting Misclassified and
  Out-of-Distribution Examples in Neural Networks},'' in \emph{Int. Conf.
  Learn. Represent.}, 2017.

\bibitem{odin2017}
S.~Liang, Y.~Li, and R.~Srikant, ``{Enhancing the Reliability of
  Out-of-Distribution Image Detection in Neural Networks},'' in \emph{Int.
  Conf. Learn. Represent.}, 2018.

\bibitem{maxlogit2019}
D.~Hendrycks, S.~Basart, M.~Mazeika, M.~Mostajabi, J.~Steinhardt, and D.~X.
  Song, ``{Scaling Out-of-Distribution Detection for Real-World Settings},'' in
  \emph{Int. Conf. Mach. Learn.}, 2022.

\bibitem{energy2020}
W.~Liu, X.~Wang, J.~Owens, and Y.~Li, ``{Energy-Based Out-of-Distribution
  Detection},'' \emph{Adv. Neural Inf. Process. Syst.}, vol.~33, pp.
  21\,464--21\,475, 2020.

\bibitem{du2022vos}
X.~Du, Z.~Wang, M.~Cai, and Y.~Li, ``{VOS: Learning What You Don’t Know by
  Virtual Outlier Synthesis},'' in \emph{Int. Conf. Learn. Represent.}, 2022.

\bibitem{Wilson_2023_ICCV}
S.~Wilson, T.~Fischer, F.~Dayoub, D.~Miller, and N.~S{\"u}nderhauf, ``{SAFE:
  Sensitivity-Aware Features for Out-of-Distribution Object Detection},'' in
  \emph{Proc. Int. Conf. Comput. Vis.}, October 2023, pp. 23\,565--23\,576.

\bibitem{wu2023deep}
A.~Wu, D.~Chen, and C.~Deng, ``{Deep Feature Deblurring Diffusion for Detecting
  Out-of-Distribution Objects},'' in \emph{Proc. IEEE Conf. Comput. Vis.
  Pattern Recog.}, 2023, pp. 13\,381--13\,391.

\bibitem{kumar2023normalizing}
N.~Kumar, S.~{\v{S}}egvi{\'c}, A.~Eslami, and S.~Gumhold, ``{Normalizing Flow
  Based Feature Synthesis for Outlier-Aware Object Detection},'' in \emph{Proc.
  IEEE Conf. Comput. Vis. Pattern Recog.}, 2023, pp. 5156--5165.

\bibitem{huang2022out}
C.~Huang, V.~Abdelzad, C.~G. Mannes, L.~Rowe, B.~Therien, R.~Salay,
  K.~Czarnecki \emph{et~al.}, ``{Out-of-Distribution Detection for LiDAR-Based
  3D Object Detection},'' in \emph{Int. Conf. Intell. Transp. Syst.}\hskip 1em
  plus 0.5em minus 0.4em\relax IEEE, 2022, pp. 4265--4271.

\bibitem{scholkopf1999support}
B.~Sch{\"o}lkopf, R.~C. Williamson, A.~Smola, J.~Shawe-Taylor, and J.~Platt,
  ``{Support Vector Method for Novelty Detection},'' \emph{Adv. Neural Inf.
  Process. Syst.}, vol.~12, 1999.

\bibitem{dinh2016density}
L.~Dinh, J.~Sohl-Dickstein, and S.~Bengio, ``{Density Estimation Using Real
  NVP},'' in \emph{Int. Conf. Learn. Represent.}, 2017.

\bibitem{caesar2020}
H.~Caesar, V.~Bankiti, A.~H. Lang, S.~Vora, V.~E. Liong, Q.~Xu, A.~Krishnan,
  Y.~Pan, G.~Baldan, and O.~Beijbom, ``{nuScenes: A Multimodal Dataset for
  Autonomous Driving},'' in \emph{Proc. IEEE Conf. Comput. Vis. Pattern
  Recog.}, 2020, pp. 11\,621--11\,631.

\bibitem{qi2017pointnet}
C.~R. Qi, H.~Su, K.~Mo, and L.~J. Guibas, ``{PointNet: Deep Learning on Point
  Sets for 3D Classification and Segmentation},'' in \emph{Proc. IEEE Conf.
  Comput. Vis. Pattern Recog.}, 2017, pp. 652--660.

\bibitem{qi2017pointnet++}
C.~R. Qi, L.~Yi, H.~Su, and L.~J. Guibas, ``{PointNet++: Deep Hierarchical
  Feature Learning on Point Sets in a Metric Space},'' \emph{Adv. Neural Inf.
  Process. Syst.}, vol.~30, 2017.

\bibitem{zhou2018voxelnet}
Y.~Zhou and O.~Tuzel, ``{VoxelNet: End-to-End Learning for Point Cloud Based 3D
  Object Detection},'' in \emph{Proc. IEEE Conf. Comput. Vis. Pattern Recog.},
  2018, pp. 4490--4499.

\bibitem{yan2018second}
Y.~Yan, Y.~Mao, and B.~Li, ``{SECOND: Sparsely Embedded Convolutional
  Detection},'' \emph{Sensors}, vol.~18, no.~10, p. 3337, 2018.

\bibitem{lang2019pointpillars}
A.~H. Lang, S.~Vora, H.~Caesar, L.~Zhou, J.~Yang, and O.~Beijbom,
  ``{PointPillars: Fast Encoders for Object Detection from Point Clouds},'' in
  \emph{Proc. IEEE Conf. Comput. Vis. Pattern Recog.}, 2019, pp.
  12\,697--12\,705.

\bibitem{yin2021center}
T.~Yin, X.~Zhou, and P.~Krahenbuhl, ``{Center-Based 3D Object Detection and
  Tracking},'' in \emph{Proc. IEEE Conf. Comput. Vis. Pattern Recog.}, 2021,
  pp. 11\,784--11\,793.

\bibitem{hendrycks2018ood}
D.~Hendrycks, M.~Mazeika, and T.~Dietterich, ``{Deep Anomaly Detection with
  Outlier Exposure},'' in \emph{Int. Conf. Learn. Represent.}, 2019.

\bibitem{hornauer2023heatmap}
J.~Hornauer and V.~Belagiannis, ``{Heatmap-Based Out-of-Distribution
  Detection},'' in \emph{Proc. IEEE Winter Conf. Appl. Comput. Vis.}, 2023, pp.
  2603--2612.

\bibitem{lee2018simple}
K.~Lee, K.~Lee, H.~Lee, and J.~Shin, ``{A Simple Unified Framework for
  Detecting Out-of-Distribution Samples and Adversarial Attacks},'' \emph{Adv.
  Neural Inf. Process. Syst.}, vol.~31, 2018.

\bibitem{piroli2023ls}
A.~Piroli, V.~Dallabetta, J.~Kopp, M.~Walessa, D.~Meissner, and K.~Dietmayer,
  ``{LS-VOS: Identifying Outliers in 3D Object Detections Using Latent Space
  Virtual Outlier Synthesis},'' in \emph{Int. Conf. Intell. Transp. Syst.},
  2023, pp. 1242--1248.

\bibitem{cen2022openworld}
J.~Cen, P.~Yun, S.~Zhang, J.~Cai, D.~Luan, M.~Tang, M.~Liu, and M.~Yu~Wang,
  ``{Open-World Semantic Segmentation for LIDAR Point Clouds},'' in \emph{Eur.
  Conf. Comput. Vis.}\hskip 1em plus 0.5em minus 0.4em\relax Springer, 2022,
  pp. 318--334.

\bibitem{he2017mask}
K.~He, G.~Gkioxari, P.~Doll{\'a}r, and R.~Girshick, ``{Mask R-CNN},'' in
  \emph{Proc. Int. Conf. Comput. Vis.}, 2017, pp. 2961--2969.

\bibitem{girshick2014rich}
R.~Girshick, J.~Donahue, T.~Darrell, and J.~Malik, ``{Rich Feature Hierarchies
  for Accurate Object Detection and Semantic Segmentation},'' in \emph{Proc.
  IEEE Conf. Comput. Vis. Pattern Recog.}, 2014, pp. 580--587.

\bibitem{dosovitskiy2017carla}
A.~Dosovitskiy, G.~Ros, F.~Codevilla, A.~Lopez, and V.~Koltun, ``{CARLA: An
  Open Urban Driving Simulator},'' in \emph{Conf. on Robot Learning}.\hskip 1em
  plus 0.5em minus 0.4em\relax PMLR, 2017, pp. 1--16.

\bibitem{paszke2019pytorch}
A.~Paszke, S.~Gross, F.~Massa, A.~Lerer, J.~Bradbury, G.~Chanan, T.~Killeen,
  Z.~Lin, N.~Gimelshein, L.~Antiga \emph{et~al.}, ``{PyTorch: An Imperative
  Style, High-Performance Deep Learning Library},'' \emph{Adv. Neural Inf.
  Process. Syst.}, vol.~32, 2019.

\bibitem{mmdet3d2020}
M.~Contributors, ``{{MMDetection3D: OpenMMLab}} next-generation platform for
  general {3D} object detection,''
  \url{https://github.com/open-mmlab/mmdetection3d}, 2020.

\bibitem{sutskever2013importance}
I.~Sutskever, J.~Martens, G.~Dahl, and G.~Hinton, ``{On the Importance of
  Initialization and Momentum in Deep Learning},'' in \emph{Int. Conf. Mach.
  Learn.}\hskip 1em plus 0.5em minus 0.4em\relax PMLR, 2013, pp. 1139--1147.

\end{thebibliography}

\end{document}